\pdfoutput=1

\documentclass[11pt]{article}
\usepackage[final]{acl} 

\usepackage{times}
\usepackage{latexsym}
\usepackage[T1]{fontenc}
\usepackage[utf8]{inputenc}
\usepackage{microtype}
\usepackage{inconsolata}

\usepackage{nicefrac}       
\usepackage{microtype}      
\usepackage{booktabs}       
\usepackage{natbib} 
\usepackage{doi}            
\usepackage{adjustbox}      
\usepackage{float}          
\usepackage{graphicx}       
\usepackage{multicol}       
\usepackage{multirow}       
\usepackage{amssymb}        
\usepackage{longtable}  
\usepackage{float} 
\usepackage{placeins}

\title{Multi-modal Preference Alignment Remedies Degradation of Visual Instruction Tuning on Language Models}

\author{Shengzhi Li \\
  TIFIN Inc \\
  \texttt{alex.li@tifin.com} \\\And
  Rongyu Lin \\
  KAUST \\
  \texttt{rongyu.lin@kaust.edu.sa} \\\And
  Shichao Pei\thanks{Corresponding author.} \\
  University of Massachusetts Boston \\
  \texttt{shichao.pei@umb.edu} \\
}
\begin{document}
\maketitle

\begin{abstract}
Multi-modal large language models (MLLMs) are expected to support multi-turn queries of interchanging image and text modalities in production. However, the current MLLMs trained with visual-question-answering (VQA) datasets could suffer from degradation, as VQA datasets lack the diversity and complexity of the original text instruction datasets with which the underlying language model was trained. To address this degradation, we first collect a lightweight, 5k-sample VQA preference dataset where answers were annotated by Gemini for five quality metrics in a granular fashion and investigate standard Supervised Fine-tuning, rejection sampling, Direct Preference Optimization (DPO) and SteerLM algorithms. Our findings indicate that with DPO, we can surpass the instruction-following capabilities of the language model, achieving a 6.73 score on MT-Bench, compared to Vicuna's 6.57 and LLaVA's 5.99. This enhancement in textual instruction-following capability correlates with boosted visual instruction performance (+4.9\% on MM-Vet, +6\% on LLaVA-Bench), with minimal alignment tax on visual knowledge benchmarks compared to the previous RLHF approach. In conclusion, we propose a distillation-based multi-modal alignment model with fine-grained annotations on a small dataset that restores and boosts MLLM's language capability after visual instruction tuning. 
\end{abstract}

\section{Introduction}
Recent advancements in artificial intelligence have led to the rise of multi-modal large language models (MLLMs), which combine textual and visual interpretation capabilities in a single model \cite{shen2023hugginggpt}. However, effectively blending multi-modality in one system has proven non-trivial. Integrating diverse data forms often creates internal representation conflicts, giving rise to the issue known as "catastrophic forgetting" \cite{kirkpatrick2017overcoming}. 
The diversity constraint in visual question answering (VQA) datasets could be attributed as a source of the issue. VQA tasks typically focus on descriptive queries about image contents, whereas textual datasets encompass a broader range of complex cognitive tasks, including reasoning, writing, summarization, and coding. This discrepancy in dataset complexity is a key factor contributing to the observed performance degradation in MLLMs. Our evaluation of models such as BLIP-2, InstructBLIP, and LLaVA against language instruction-following benchmarks like MT-Bench \cite{zheng2023judging} and AlpacaEval \cite{alpaca_eval} revealed diminished language capabilities in comparison to their linguistic backbones. For instance, LLaVA, built on the Vicuna-13b LLM, demonstrated a decline in MT-Bench performance from 6.57 to 5.92, even underperforming the Vicuna-7B model.

Driven by the limitations observed in distillation-based instruction tuning, particularly its constrained generalizability and the narrow performance improvements on tasks outside the training distribution, this study investigates the efficacy of distillation-based preference alignment in addressing modality conflict in MLLMs. The decision to explore this avenue is predicated on the hypothesis that integrating AI-generated preference data can provide a more granular and nuanced alignment with human expectations, potentially mitigating the adverse effects of modality conflict. 

This study rigorously evaluates three baseline methodologies—Direct Preference Optimization (DPO), SteerLM, and Rejection Sampling—as potential solutions to utilize the distilled preference data to enhance the instruction-following capabilities and address the modality conflict inherent in MLLMs. Each of these methods offers a unique approach to model alignment, from the direct optimization of preferences in DPO to the conditional supervision in SteerLM and the selective acceptance in Rejection Sampling. Our empirical analysis reveals that DPO, in particular, demonstrates a pronounced efficacy in reconciling the performance disparities observed between textual and visual modalities. By leveraging a refined preference dataset, fine-tuned with the DPO objective and supplemented with comprehensive annotations from advanced AI models, DPO not only addresses the modality conflict but also significantly enhances the model's performance across a spectrum of benchmarks. The results indicate that, through the application of DPO, MLLMs can achieve a more robust alignment with human-like preferences, thereby mitigating the adverse effects of catastrophic forgetting and modality conflict and elevating the models' capabilities to a level that surpasses traditional instruction tuning methods.

Our main contributions are: 

\textbf{Exploration of Modality Degradation:} This work is the first to identify and address modality degradation in MLLMs, a phenomenon where visual instruction tuning detrimentally impacts language instruction capabilities. Our systematic investigation into this issue contributes novel insights to the field, laying the groundwork for further research in mitigating such degradation.

\textbf{Efficient and scalable preference alignment pipeline as remedy} Our data collection strategy employs a granular quality metric annotation format, leveraging cost-effective commercial APIs. This scalable approach enables the efficient production of high-quality datasets. We are able to surpass LLaVA and Vicuna's language instruction-following capability with DPO on a 6k dataset. 

\section{Related Work}

\subsection{MLLMs and Visual Instruction Tuning} 
Incorporating another modality into large language models represents a natural evolution for these systems. Modality expansion can be achieved through system-level enhancements at inference time, with approaches such as Mm-react \cite{yang2023mm}, Visual ChatGPT \cite{wu2023visual}, and HuggingGPT \cite{shen2023hugginggpt} enabling the LLM to invoke off-the-shelf vision models and APIs. An alternative strand of research involves the training of end-to-end MLLMs. To avoid the prohibitive costs associated with pre-training from scratch, these models often integrate pre-trained vision models with large language models, applying various degrees of modality adaptation. Mini-GPT4 \cite{gong2023multimodal} focuses solely on training a linear projection matrix to connect CLIP-based \cite{radford2021learning} vision representations with the LLaMA model \cite{touvron2023llama}; BLIP-2 introduces a cross-attention module to extract vision tokens relevant to the query. Both LLaVA \cite{liu2023visual} and mPlug-OWL \cite{ye2023} feature cross-modality connectors between the vision and language domains, but they also fine-tune the LLM and vision encoder, respectively. Flamingo \cite{alayrac2022flamingo}, in contrast, incorporates new cross-attention layers directly into the LLM.

In the language domain, \citet{wei2021finetuned} discovered that fine-tuning a base LLM with instructions described in natural language enhances the model's ability to follow those instructions. In a similar vein, MLLMs are typically fine-tuned with visual instructions; Mini-GPT4 \cite{gong2023multimodal} utilized template instructions based on image-text pairs, while InstructBLIP \cite{dai2023instructblip}, Otter \cite{li2023otter}, and LLaVA \cite{liu2023visual} employed human-written visual question-answers or synthetically generated question-answer pairs by prompting GPT-4 with COCO captions and bounding boxes \cite{liu2023visual}. However, considering that both LLaVA and Instruct-BLIP utilize Vicuna \cite{vicuna2023}, it remains a topic of debate whether their steps of visual instruction tuning genuinely add to the model's instruction-following capabilities or merely conform to the instruction-following format.

\subsection{Mitigating Modality Conflict 
in MLLMs}
In response to modality conflicts, LLaVA \cite{liu2023visual} and mPLUG-OWL \cite{ye2023mplugowl2} reincorporate language instruction data from ShareGPT into their training datasets. Despite this, LLaVA 1.5's language instruction-following degrades significantly. As measured by MT-Bench, LLaVA-1.5-13B lags behind its Vicuna-V1.5-13b (5.99 vs 6.57) or even the smaller Vicuna-V1.5-7b (5.99 vs 6.17). While mPLUG-OWL-2 shows promise on text benchmarks and addresses modality conflicts with new architecture features, it does so at the cost of increased complexity and a higher parameter count, rising from 7.2 billion to 8.2 billion. These modifications necessitate greater computational resources and complicate the inference process due to separate processing pathways for different modalities.

\subsection{Distillation-based Instruction Tuning} 


Smaller open-source models, including Vicuna \cite{vicuna2023}, Alpaca, and more recently ShareGPT4V \cite{chen2023sharegpt4v} have been fine-tuned using the outputs of large proprietary models, yet this approach often falls short in terms of broadening the models' generalization capabilities. \citet{gudibande2023false} criticizes the reliance on imitation for instruction tuning, noting that it fails to substantially bridge the performance gap for tasks beyond the training data's scope. Our findings suggest that such strategies do not effectively extend to out-of-distribution scenarios. In fact, small-scale distillation efforts tend to overly specialize the model, diminishing its performance on general benchmarks. This issue was highlighted in an experiment where fine-tuning LLaVA with a 5k VQA dataset from Gemini Pro-generated answers led to notable declines across textual and visual benchmarks. Our research takes a novel approach by exploring alignment-based methods such as Direct Preference Optimization (DPO), which not only preserves but enhances the model's capabilities across diverse benchmarks, thereby offering a more effective solution to the limitations of traditional distillation-based instruction tuning.
\subsection{Preference Alignment}
The Instruct-GPT series \cite{ouyang2022training} has shown that merely employing supervised fine-tuning (SFT) on Large Language Models (LLMs) is insufficient for aligning them with human preferences. The technique of Reinforcement Learning from Human Feedback (RLHF) \cite{ouyang2022training} addresses this by constructing a reward model that encapsulates human preferences and then applying reinforcement learning to maximize this reward. The Direct Preference Optimization (DPO) approach posits that directly tuning the preference dataset can serve as an effective substitute for reward modeling, offering the added benefit of reduced computational complexity. SteerLM has recently been identified to achieve performance akin to RLHF by incorporating human-annotated quality metrics before generation, serving as a conditional SFT-based strategy for alignment \cite{dong2023steerlm}. Rejection sampling selects the highest-rated self-generated sampling and applied SFT and has been used in combination with RLHF in works such as LLaMA-2 \cite{touvron2023}. Our experiments with DPO, SteerLM, and rejection sampling are compared to the prior work applying RLHF to LLaVA models as a baseline. 

\subsection{Distilling AI Feedback for Preference Alignment}
In the realm of alignment methods, reliance on human-annotated preference annotations is common. While effective on a large scale, this approach incurs substantial costs and operational complexities \cite{touvron2023}. The effectiveness of reward models based on pairwise ranking is constrained by the inherent subjectivity of human preferences, with LLaMA's reward model achieving an accuracy range of 64.3-70.6\% and the LLaVA-RLHF model reaching 67\%. In response to these limitations, Zephyr \cite{tunstall2023zephyr} and UltraFeedback \cite{cui2023ultrafeedback} have utilized preference annotations distilled from GPT-4 to train models with 7B parameters and achieve promising performance. Motivated to adapt this distillation-preference alignment approach for MLLMs, our work introduces a chain-of-thought prompting coupled with a granular annotation guide spanning five assessment metrics.

\begin{figure*}[t]
    \centering
    \label{tag:MethodIllustration}
    \includegraphics[width=1\linewidth]{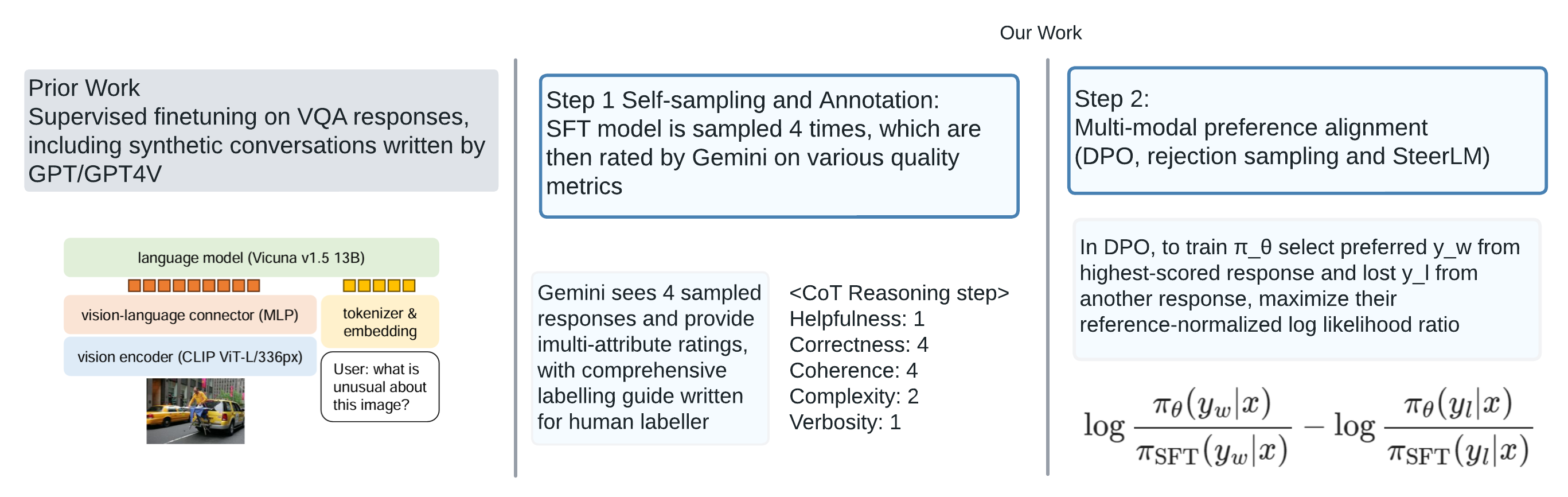}
    \caption{From a visual-instruction-tuned pre-trained model, we generate 4 completions for a given image-question prompt. These answers are then presented to Gemini to obtain granular annotations given a labeling guide. We construct a preference dataset of (image-text prompt, preferred completion) and (image-text prompt, rejected completion). We benchmarked DPO, Rejection sampling, and SteerLM alignment methods, in addition to a pure SFT baseline using Gemini-provided answers directly. }

    \label{tag:MethodIllustration}

\end{figure*}
\begin{table*}[t]
  \centering
  \begin{minipage}{\textwidth}  
    \centering
    \small
    \renewcommand{\arraystretch}{1.3}  
    \caption{Granular annotation format and labelling guide proposed in HelpSteer \cite{Wang2023Nov} used in VQA annotation collection. LLaVA-1.5-13b generates 4 candidates. Using the HelpSteer labeling guide, images, and questions, Gemini rates each completion with a score of 0-4 in each metric.}
    \begin{tabular}{|l|p{0.7\textwidth}|}  
      \hline
      \textbf{Metric} & \textbf{Description} \\
      \hline
      Helpfulness & Measures if the response fully addresses the prompt's request. \\
      \hline
      Correctness & Assesses the accuracy and relevance of the information, ensuring it's free from errors or misinformation. \\
      \hline
      Coherence & Evaluates the clarity and logical consistency of the response. \\
      \hline
      Complexity & Looks at the level of sophistication in the language used, from simple to advanced. \\
      \hline
      Verbosity & Considers the brevity or lengthiness of the response in relation to the prompt's needs. \\
      \hline
    \end{tabular}
    \label{fig:annotation-guide}
  \end{minipage}
\end{table*}

\section{Method}
The focus of this study is to investigate improving the text instruction following of MLLMs while retaining or potentially enhancing their multi-modal capabilities. 
For this purpose, we propose to harness alignment methods that utilize self-sampled responses and preference annotations in addition to Supervised Fine-Tuning (SFT) as a baseline.

\subsection{Data Collection}
 

\textbf{Sources of Multi-modal Prompts.} We have curated 5,000 samples from SciGraphQA \cite{Li2023Aug} and  LRV-Instruct \cite{liu2023aligning} to assemble the image-text prompts for our multi-modal preference dataset. SciGraphQA is a multi-turn question-answering dataset regarding scientific graphs. The LRV-Instruct dataset is a visual instruction dataset aimed at mitigating hallucinations by incorporating both positive prompts (inquiring about objects present in the image) and negative prompts (requesting information about absent objects) \cite{liu2023aligning}. Including negative examples encourages the multi-modal LLM to critically evaluate the prompts and identify instances where the requested objects are absent in the image. The LRV-Instruct dataset enhanced its performance in reducing hallucinations, as evidenced by its application in the Pope framework \cite{liu2023aligning}.

\begin{table}[t]
\centering
\label{table:modelparamters}
\caption{Data mixture: We applied about 2.5K from the LRV-Instruct, an instruction tuning dataset aiming at addressing hallucination with 'negative examples' referring to nonexistent objects, and SciGraphQA \cite{Li2023Aug}, a multi-turn question-answer dataset regarding scientific graphs. Our SFT method relied on answers from the respective datasets directly, while DPO, SteerLM, and rejection sampling methods use self-generated instead.}
\begin{tabular}{lll}
\hline
\textbf{Data Type} & \textbf{Data Name} & \textbf{Size} \\ \hline
VQA & LRV-Instruct & 2562\\
 & SciGraphQA & 2522\\
Total & & 5084\\ \hline
\end{tabular}\
\end{table}
\textbf{Granular Annotation}: We generate 4 completions for the dataset above using LLaVA-1.5-13B with a temperature of 0.7 and then prompt Gemini-Vision \cite{team2023gemini} with the labeling guide of HelpSteer gave to Scale AI workers \cite{Wang2023Nov}, images, questions, and the 4 completions, obtaining granular annotations for various quality metrics, including helpfulness, correctness, and coherence, providing multifaceted insights into the model's outputs.  We leveraged Zero-Shot Chain-of Thought prompting \cite{kojima2022large} such that the Gemini gives the reasoning for rating each response, an inner calibration monologue, and ratings as shown in \ref{tag:MethodIllustration}. The appendix provides two annotation examples from the LLaVA-RLHF dataset \cite{sun2023aligning} to visualize how Gemini reasons and rates each metric compared to binary preference annotations provided by crowd-sourced workers.

We selected Gemini Pro (dated 01/01/2024) for its performance, cost efficiency, and bias mitigation \cite{team2023gemini}. Leading the OpenCompass multi-modal leaderboard with an average rank of 1.89, outperforming GPT-4V's 2.89 \cite{opencompass}, its generous free tier supports extensive data collection. Crucially, as GPT-4 is used as a judge in multiple instruction-following benchmarks, using GPT-4(V) for both benchmark judging and data labeling could introduce bias, potentially skewing our models towards GPT-4's preferences. This consideration led us to opt for Gemini to ensure a more objective evaluation of our fine-tuning efforts.

\subsection{Alignment Methods}

\textbf{Self-sampled SteerLM}: SteerLM, a conditional Supervised Fine-Tuning (SFT) technique, aligns LLMs by augmenting prompts with a description of the desired response quality, as introduced by \cite{dong2023steerlm}. This method conditions SFT on granular annotation generated by Gemini and surpasses traditional SFT and alignment strategies. We construct a conditional prompt template for this technique by incorporating a conditional prompt following HelpSteer guidelines. For instance, 

\noindent
\textbf{Rejection Sampling}: Following Constitutional AI \cite{bai2022constitutional} and LLaMA-2 \cite{touvron2023}, we adopt a simplified rejection sampling approach. Specifically, we select the top-scoring response from the four completions described above based on Gemini's aggregated scores for helpfulness, correctness, and coherence. We apply standard SFT based on the selected responses without additional prompt conditions, following a straightforward format: (image, prompt, best response).

\noindent
\textbf{Direct Preference Optimization (DPO)}: In the DPO approach, we first convert our dataset of granular annotations into a preference set. This conversion involves evaluating four responses per sample, each scored by Gemini on metrics including Helpfulness and Correctness. To form the preference set, we aggregate these scores to determine the overall quality of each response. The response with the highest aggregated score is designated as the 'preferred' response. The rejected response is selected from the remaining options. To qualify as a 'rejected' response, it must have a summed score that is at least 2 points lower than that of the preferred response. This selection process filters out pairs of preferred and rejected responses that could not be clearly distinguished. This method allows us to harness robust preference signals that enhance the training of the DPO model, ensuring that it aligns more closely with granular quality signals. We follow the method outlined in \cite{rafailov2023direct} for implementing the loss function, which effectively utilizes these preference pairs to optimize the model's performance.

\noindent
\textbf{SFT from Gemini Responses} This is an important baseline not using self-sampling but using answers from Gemini directly for pure SFT. Gemini was prompted with the questions and images from our dataset. We gather the answer directly from Gemini answering the question and use the same hyper-parameters in SteerLM and Rejection Sampling. 

\section{Experiments}

\subsection{Experimental Setup}
For training, we utilize Azure Cloud (NC-A100 series) with 4 A100-80G GPUs. In all experiments, LoRA is employed for parameter-efficient tuning. For the SFT experiments, including standard SFT with Gemini responses, rejection sampling, and self-sampled SteerLM, we adhered to the same hyperparameters used in LLaVA-v1.5's instruction tuning. 

We conducted a hyperparameter search to optimize the performance of the DPO model on the LLaVA bench, an open-ended instruction tuning benchmark. Our experiment involved 1,000 samples to evaluate the influence of the beta term, which we hypothesized to significantly determine the divergence of the finetuned DPO model from the original model, potentially leading to knowledge catastrophic forgetting. The results are in the Appendix \ref{tab:DPO-hyperparam}

Based on these results, we selected the optimal hyperparameter set: Beta of 0.1, not using average log probability and a learning rate of 5e-5. This configuration led to the highest LLaVA-bench score of 77.4, indicating superior performance in instruction tuning. The LLaVA bench, while primarily an instruction tuning benchmark, does not exclusively focus on knowledge. Future work will need to consider whether to fit a knowledge benchmark exclusively or a mix of knowledge and instruction tuning benchmarks to further validate the robustness and applicability of the DPO model.

\begin{table*}[t]
\centering
\renewcommand\tabcolsep{20pt}
\small
\label{table}
\caption{Model parameters and hyperparameters setup for reproduction. The base model is LLaVA-v1.5-13b, derived from Vicuna-1.5-13b (LLama-2-13b). The training was conducted using Deepspeed Stage-3 on a 4x A100 80GB GPU machine with LoRA for parameter-efficient fine-tuning. DPO and SFT (including SteerLM and rejection sampling) employed distinct hyperparameter sets optimized for performance.}
\begin{tabular}{lll}
\hline
\textbf{Parameter Settings} & \textbf{Name} & \textbf{Value} \\ \hline
Model architecture & Pre-trained model & LLaVA-v1.5-13b \\
& Language Backbone & Vicuna-v1.5-13B  \\
& Vision Tower & clip-vit-large \\
& Multi-modal Adapter &  2-layer Multi-layer-Perception \\
Lora Setting &  Lora Rank & 128 \\
&Lora Alpha & 256 \\
DPO Setting & Average Log Probability & False \\
& Fine-grained DPO & False \\
&Beta & 0.1 \\
&Learning Rate & 5e-5 \\
& Gradient Accumulation Steps & 4 \\
& Per device train batch size &8 \\
&Model Max length & 3000 \\
SFT Setting & Learning Rate & 4e-4 \\
& Per device train batch size &16 \\
& Model Max Length  & 2048 \\
Common settings & Hardware & 4X A100 80G \\
& Distributed Learning & Zero-3 \\
& Use BF-16 & True \\
& Learning Rate Scheduler & Cosine \\
& Learning Rate Warm up  & 0.003 \\
& Weight Decay  & False \\\hline
\end{tabular}
\label{tab:Modelparameters}
\end{table*}
In our DPO experiments, we performed a hyperparameter search based on 1000 samples on LLaVA Bench, exploring various values for beta (0.1, 0.2, and 0.3, averaged and non-averaged log probabilities, and learning rates of 5e-5 and 5e-6. We pre-computed the log probabilities of the reference model (LLaVA-v1.5-13b) for our preference dataset before training. The complete list is provided in the Appendix. 

\subsection{Benchmarks}
We have the following three sets of benchmarks. \subsection*{Visual Instruction Benchmarks}
\begin{itemize}
    \item \textbf{MM-Vet}: MM-Vet measures six core visual-language capabilities across 128 tasks, providing a comprehensive evaluation of multi-modal understanding. It blends math, reasoning, and visual knowledge. Each task is scored by GPT-4 based on the correctness of the model's response relative to GPT-4-generated ground truth, with scores ranging from 0 to 1, aggregated across 187 samples \cite{yu2023mmvet}.
    \item \textbf{LLaVA-Bench}: LLaVA-Bench (in the wild) comprises 60 tasks designed to test visual instruction-following and question-answering capabilities in natural settings. Scoring follows the same methodology as MM-Vet, where GPT-4 generates the ground truth answers and scores the model's responses from 0 to 1 \cite{liu2023visual}.
\end{itemize}

\subsection*{Visual Multi-Choice Benchmarks}
\begin{itemize}
    \item \textbf{PoPE}: An object hallucination benchmark with 8,440 samples aimed at evaluating the model's ability to accurately discern and describe visual content. This benchmark uses yes-no binary classification evaluations, reporting F-1 scores, and questions are structured to probe the model on the presence of objects within images, such as ‘Is there a car in the image’ \cite{li2023evaluating}.
    \item \textbf{MM-Bench}: A comprehensive multi-modal benchmark with 4.7K samples featuring multiple-choice visual knowledge and reasoning tasks. To prevent bias towards any particular answer choice, questions are assessed through rotational augmentations (e.g., ABC, BAC), and a question is considered correctly answered only if all rotational formats are correct. Overall success is reported as a total accuracy percentage \cite{liu2023mmbench}.
\end{itemize}

\subsection*{Language Instruction-Following Benchmarks}
\begin{itemize}
    \item \textbf{MT-Bench}: Utilizes LLMs to approximate human preferences with over 80\% agreement, focusing on measuring the helpfulness of responses across 160 samples in single-turn and multi-turn settings. The benchmark evaluates responses based on their alignment with GPT-4-generated preferred responses \cite{zheng2023judging}.
    \item \textbf{AlpacaEval}: Leverages GPT-4 to assess the percentage of cases where the candidate LLM outperforms the GPT-3 API (text-davinci-003) across 160 evaluations. This benchmark highlights advancements in LLM capabilities over previous model iterations \cite{alpaca_eval}.
\end{itemize}

\begin{table*}[t]
\small
\caption{Performance comparison among alignment strategies. The results demonstrate DPO-13B's superior performance, particularly in reconciling language capabilities while enhancing visual task performance, validating the DPO methodology's efficacy in multi-modal alignment.}
\label{comparisiontollava-llavarlhf}
\setlength{\tabcolsep}{3pt} 
\begin{tabular}{p{3cm}|p{1.8cm}|p{2cm}|p{1.5cm}|p{2.5cm}|p{2cm}|p{1.5cm}}
\hline
\textbf{Model Name} & \multicolumn{2}{c|}{\textbf{Visual Instruction Benchmark}} & \multicolumn{2}{c|}{\textbf{Visual Multi-Choice Benchmark}} & \multicolumn{2}{c}{\textbf{Text Instruction Benchmark}} \\ \hline
& \textbf{MM-Vet} & \textbf{LLaVA-bench} & \textbf{PoPe} & \textbf{MM-Bench} & \textbf{MT-bench} & \textbf{AlpacaEval} \\ \hline
Vicuna-1.5-13b \cite{vicuna2023}& - & - & - & - & 6.57 & 81.4 \\ \hline
LLaVA-1.5-13b \cite{liu2023visual}& 36.3 & 73.1 & 0.859 & 67.4 & 5.99 & 79.3 \\ \hline
LLaVA-RLHF-13b \cite{sun2023aligning}& 37.2 & 76.8 & 0.869 & 60.1 & 6.18 & 81.0 \\ \hline
\multicolumn{7}{c}{\bf Alignment method we benchmarked, finetuning LLaVA-1.5-13b} \\ \hline
Standard SFT & 36.5 & 63.7 & 0.850 & 65.4 & 5.01 & 50.2 \\ \hline
SteerLM & 35.2 & 67.0 & 0.878 & 65.1 & 5.70 & 68.8 \\ \hline
Rejection-sampling & 38.0 & 70.6 & \textbf{0.883} & \textbf{67.6} & 6.22 & 74.9 \\ \hline
DPO & \textbf{41.2} & \textbf{79.1} & 0.870 & 66.8 & \textbf{6.73} & \textbf{86.4} \\ \hline
\end{tabular}
\end{table*}

We run benchmark code once on the target model, with greedy decoding by default. In case a benchmark uses a different decoding setting officially, we follow the benchmark's settings. 
\subsection{Results}
Table \ref{comparisiontollava-llavarlhf} illustrates a comparative analysis of various alignment methods, including RLHF, Standard SFT, SteerLM, Rejection Sampling, and DPO, aimed at enhancing the language capabilities of the LLaVA model that were compromised post-visual instruction tuning. Vicuna, the base language model of LLaVA, is tested on language tasks. 
After visual instruction tuning, LLaVA experiences a decline in language benchmark scores from Vicuna (from 6.57 to 5.99 on MT-Bench and from 81.4 to 79.3 on AlpacaEval). However, DPO, among the alignment strategies on the 5k multi-modal dataset, not only mitigates the degradation problem but also surpasses Vicuna's performance on both benchmarks. 

In open-ended visual instruction tasks, DPO significantly outperforms the baseline LLaVA and LLaVA-RLHF models on both the MM-Vet and LLaVA-bench. These open-ended benchmarks, structured similarly to MT-Bench where GPT-4 assesses the responses against a gold standard, revealed a relatively high 0.73 Pearson Correlation between MM-Vet and MT-Bench (n=7, p=0.1). 

In visual multi-choice benchmarks, PoPe, which evaluates for hallucination, and MM-Bench, which assesses world knowledge and reasoning, rejection sampling emerged as the most effective method, whereas DPO showed lesser efficacy. While DPO improved open-ended benchmarks, it slightly decreased LLaVA's MM-Bench score from 67.4 to 66.8, contrasting with LLaVA-RLHF's notable drop from 67.4 to 60.1, indicating a less significant alignment tax.
\begin{figure}[t]
    \centering
\includegraphics[width=1\linewidth]{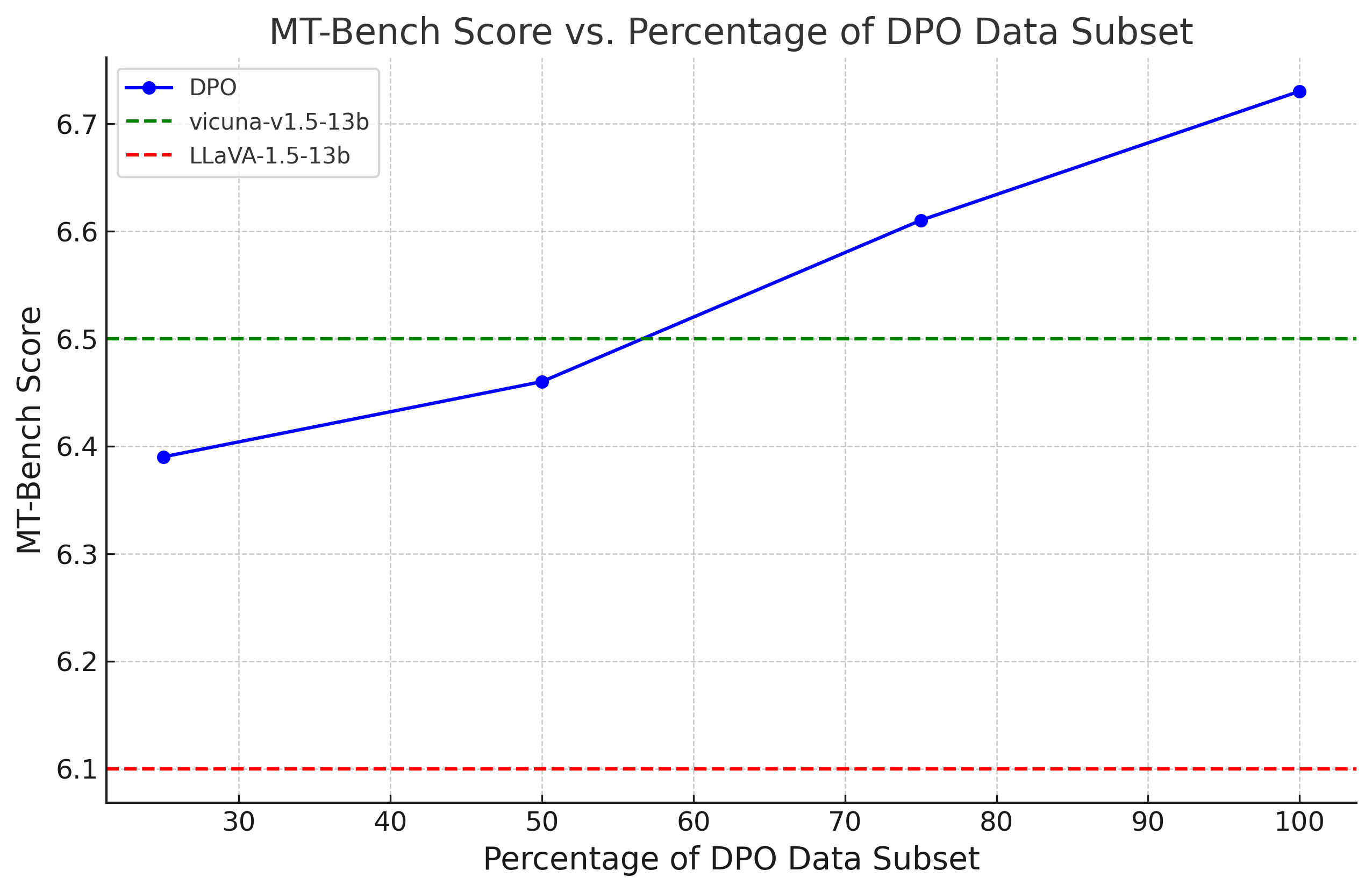}
    \caption{Advances in MT-Bench scores via DPO data scaling}

    \label{fig:datascaling}
\end{figure}
Figure \ref{fig:datascaling} illustrates the effect of scaling the DPO dataset on the MT-Bench scores, signifying the efficacy of DPO in advancing the performance of the LLaVA-v1.5-13b model after visual instruction tuning. The baseline at 0\% DPO data, marked at a score of 5.99, represents the initial performance of LLaVA-v1.5-13b. As the DPO dataset's size increases, a notable rise in MT-Bench scores is observed, peaking at 6.73 with 100\% data utilization. Remarkably, the performance surpasses that of the Vicuna-v1.5-13b benchmark using less than 75\% or 4.2K of the DPO data, underlining DPO's data efficiency. This data scaling trend emphasizes DPO's potential as an effective alignment strategy for MLLMs, addressing the challenge of performance degradation due to visual instruction tuning.

\section{Discussion}
\subsection{Limitation of collecting multi-modal preference data manually}
 The scarcity, high cost, and inconsistency of existing multi-modal preference datasets present significant challenges to model alignment. Our manual labeling of a selected subset from the LLaVA-RLHF human preference dataset underscored the difficulty in achieving consensus or identifying clear preferences, shedding light on the inherent subjectivity of these datasets. To quantify these observations, we embarked on a targeted experiment involving 500 samples. Gemini annotated two distinct responses from the dataset employed in the LLaVa-RLHF reward model's data collection \cite{sun2023aligning}. This facilitated a direct comparison between Gemini's annotations and human preferences. The correlation heatmap depicted in Figure \ref{fig:correlation-heatmap} reveals a notably weak correlation between human preferences from LLaVA-RLHF and Gemini scores, with correlation coefficients falling below 0.1. This observation was corroborated by further manual inspections, which frequently resulted in an inability to definitively determine clear preferences. This finding underscores the high subjectivity and individual bias within the LLaVA-RLHF preference data. 

\begin{table}[t]
\centering
\caption{Impact of data noising on model performance.}
\label{tab:data-noising}
\begin{tabular}{lc}
\toprule
\textbf{Data Noising Level} & \textbf{MT-Bench Score} \\
\midrule
No flip (baseline) & 6.73 \\
25\% flip & 6.35 \\
50\% flip & 6.26 \\
75\% flip & 5.99 \\
\bottomrule
\end{tabular}
\end{table}
\subsection{Empirical validation of data quality}
To further validate the quality and reliability of our annotations, we conducted a data noising experiment by randomly flipping the polarity of (preferred, rejected) pairs in the annotations and measured the trained model's MT-Bench score \ref{tab:data-noising}. We hypothesized that if our annotations contained significant noise, flipping the polarity would not substantially affect the model's performance. However, the results of the experiment indicate otherwise, with a clear trend showing performance degradation as the level of data noising increased. These findings suggest that the original annotations possess a strong and consistent signal. The model’s performance degradation with increased noise levels substantiates the reliability of our annotation process with Gemini, affirming that the annotations effectively capture meaningful distinctions between preferred and rejected responses. 
\subsection{Robustness of the model with noisy context}
We introduced a variant of the language instruction-following capability benchmark MT-Bench by sending a random image sampled from the CoCo-2017 train set to simulate real-world use cases. we also note that as \ref{tag:noisymt-bench} showed, our DPO model is more robust than other baselines in real-world conditions where the user may have an irrelevant image in the context.

\subsection{Cross-model transfer ability: Can preference dataset generated by one model be transferred to other models? } To investigate if the preference alignment dataset needs to be sampled and fine-tuned on the same model, we investigated applying preferences sampled from LLaVA-13b to fine-tune a smaller model, LLaVA-7b. Notably, this approach yielded significant enhancements, as demonstrated by the performance improvements on key benchmarks. The following table encapsulates the performance metrics observed in these experiments: This cross-model transfer ability not only corroborates the versatility of our proposed approach but also implies a scalable pathway for improving multi-modal language models without necessitating the construction of new preference data for each model iteration.

\begin{table}[t]
\centering
\caption{Performance improvements with multi-modal preference data application.}
\label{tab:cross-model-transfer}
\footnotesize 
\begin{tabular}{lcc}
\toprule
\textbf{Model Name} & \textbf{MT-Bench Score} & \textbf{MM-Vet Score} \\
\midrule
Vicuna-7B-v1.5 & 6.17 & N/A \\
LLaVA-V1.5-7b  & 5.87 & 30.5 \\
DPO-7b (ours) & 6.228 & 39.8 \\
\bottomrule
\end{tabular}
\end{table}

\subsection{Multi-modal preference alignment as a data-efficient remedy to instruction tuning capabilities}
The superficial alignment hypothesis states that a model's knowledge capabilities are largely acquired during its initial pre-training stage \cite{Zhou2023May}. A corollary of this hypothesis is that alignment tuning refines the model output generation with a preferred response format rather than knowledge acquisition. As a result, models can be effectively realigned post-visual instruction using a relatively small set of examples \cite{Kirstain2021Oct}. This principle applies to MLLMs as well, which acquire multi-modal knowledge representation via visual instruction tuning \cite{liu2023visual}. However, existing work mixed large-scale text instruction data (518K out of 1.23 million in the case of mPlug-OWL 2 and 40K in the case of LLaVA-1.5). We hypothesize that the data inefficiency above is attributed to the underlying alignment strategy and demonstrate that one would need only a small alignment dataset so long as a proper alignment strategy such as DPO is utilized.


As suggested by Table \ref{comparisiontollava-llavarlhf}, Direct Preference Optimization (DPO) emerges as a computationally efficient solution for enhancing model performance in the mixed-modal alignment space. Unlike the mixing text instruction as described above or LLaVA-RLHF, which used a large 82K dataset and complex training pipeline involving reward modeling and PPO, DPO achieves significant improvements in language capabilities with a smaller dataset and one-stop training setup. A notable advantage of DPO is its minimal alignment tax, which curtails the degradation of existing knowledge, as evidenced by its performance on benchmarks like MM-Bench, where DPO shows minimal impact. This method not only enables effective alignment of multi-modal models post-visual instruction tuning but also ensures the preservation of model performance. 
\begin{table}[t]
\centering
\renewcommand\tabcolsep{4pt}
\caption{This benchmark adds an irrelevant image in context when benchmarking MT-Bench, testing for robustness in a real-world condition. Vicuna-7B and Vicuna-13B are baseline references.}
\label{tag:noisymt-bench}
\footnotesize
\begin{tabular}{lcc}
\toprule
\textbf{Model} & \textbf{LLM} & \textbf{Noisy-image MT-Bench} \\
\midrule
Vicuna 13B v1.5 & Vicuna$_{13B}$ & 6.57 \\
Vicuna 7B v1.5  & Vicuna$_{7B}$ & 6.17 \\
BLIP-2          & FlanT5 & 1.93 \\
InstructBLIP    & Vicuna$_{7B}$ & 4.73 \\
LLaVA-v1.5-13b  & Vicuna$_{13B}$ & 5.92 \\
DPO (ours) & Vicuna$_{13B}$ & 6.63 \\
\bottomrule
\end{tabular}
\label{tab:noisy-mtbench}
\end{table}
Our methodology exhibits notable proficiency in value alignment and data efficiency, yet it is imperative to acknowledge certain limitations and potential risks. One key consideration is the scalability of our approach. While our data scaling analysis suggests significant improvements up to a 6K preference dataset, the full extent of scalability beyond this threshold remains unexplored. As the foundational open-sourced models like LLaVA evolve in complexity and size, the effectiveness of our distillation-based approach might encounter diminishing returns. 

\section{Limitations}

This research, while pioneering advancements in Direct Preference Optimization (DPO) and multi-modal large language models (MLLMs), confronts several limitations. The preference dataset, derived from artificial annotations, may not fully capture the nuanced spectrum of human values, potentially affecting the model's alignment with true human preferences and exacerbating existing biases in automated decision-making. Additionally, the generalizability of these findings across different multi-modal contexts and languages is yet to be thoroughly tested, and the deployment of such models in critical domains raises significant safety risks due to the potential unreliability of responses under varied real-world conditions. These limitations highlight the need for cautious application, further empirical validation, and ongoing research to mitigate ethical concerns and enhance model robustness and fairness.

\section{Conclusion}
In this paper, we investigated and addressed the performance decline of the widely-used multi-modal LLM, LLaVA, on language instruction-following benchmarks. We investigated various alignment strategies to enhance the model's language instruction capabilities, including standard SFT, SteerLM, and DPO. Our comprehensive benchmark evaluations revealed that DPO stands out as the most effective method, not only restoring but also surpassing the language capabilities of LLaVA's language backbone (Vicuna) on both MT-Bench and AlpacaEval. Specifically, LLaVA's performance on MT-Bench had dipped from Vicuna's 6.57 to 5.99, whereas our DPO model advanced to 6.73. Beyond language proficiency, DPO also bolstered performance on multi-modal benchmarks, enhancing accuracy by 4.9\% on MM-Vet and 6\% on LLaVA-Bench. Remarkably, DPO achieved this with just 5k preference examples, proving more data-efficient than other strategies like RLHF, which typically demand over 80k training examples. Our findings highlight DPO's minimal alignment tax effectively boosting one performance area without substantially detracting from another. This aspect is particularly valuable for practitioners aiming to maintain the pre-training knowledge and avoid catastrophic forgetting.

\bibliography{acl_latex}
\appendix

\section{Appendix}
\label{appendix}
\clearpage

\begin{table}
\caption{Examples of Annotation}
  \begin{minipage}{0.99\textwidth}
\centering  
\vspace{-4mm}
\scalebox{0.88}{
\begin{tabular}{l p{5.4cm} p{7.1cm} }
\toprule

 \multicolumn{3}{l}{\bf Gemini-Annotation example}  \\
\midrule
&  \includegraphics[height=4.5cm]{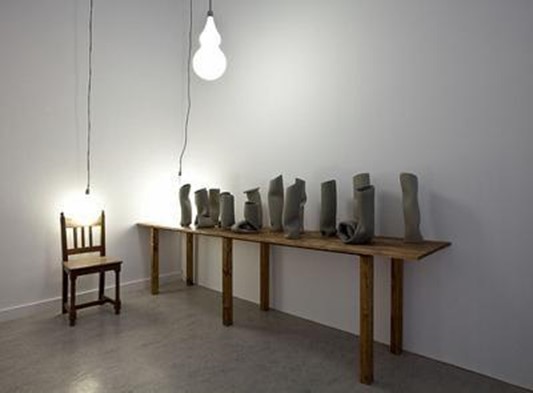} &  \includegraphics[height=4.5cm]{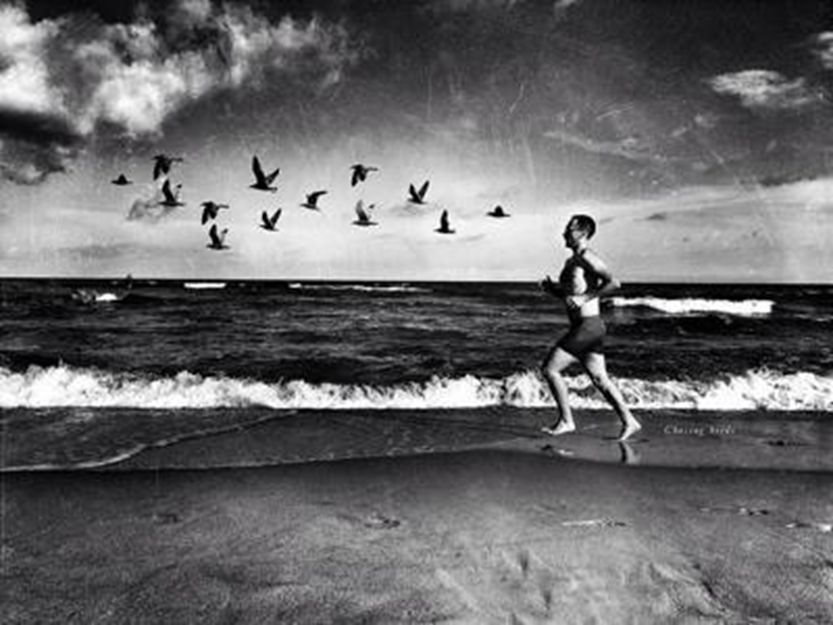} \\
Question & What aspects of the room might have artistic or creative significance?  & What do you see happening in this image?
\\ \midrule
Response 1 & In this image, the room has some artistic or creative significance due to the presence of several sculptures placed on a long wooden table. The sculptures appear to be made from various materials, such as stone and wood. Additionally, the room features a chair and a light fixture that resembles a boot. These elements, combined with the assortment of sculptures, give the room an eclectic and artistic atmosphere, suggesting that the space might be used for displaying art, inspiration, or as a workspace for an artist or designer. The presence of the boot-shaped light also adds a unique and imaginative touch to the room's overall aesthetic. & The image features a person running along a beach near the water's edge. Behind the person, multiple birds are flying in the sky, creating a dynamic and lively scene. The man is wearing a swimsuit and appears to be enjoying his time at the beach. The birds' positions vary across the sky, with some flying closer to the person and others further away, spreading out across the background.
\\ \midrule
Gemini-Rating for Response 1  & Helpfulness:4, Correctness 4, Coherence: 4, Complexity: 3, Verbosity: 3. & Helpfulness:3, Correctness 4, Coherence: 4, Complexity: 2, Verbose: 2
\\ \midrule
Response 2  & In the room, a collection of wooden shoes on a long table next to a chair can be considered as having artistic or creative significance. The display of shoes may represent a unique form of artistic expression or a design concept. Additionally, the white-walled setting with a dim light allows for a minimalist and visually striking atmosphere that enhances the shoes' display and creates a sense of depth and artistic interest in the space. & The image captures a beach scene where a man is running along the shore near the water, with waves crashing behind him. The man is in motion, as he is probably enjoying an invigorating run on the beach. In the sky above, several birds fly in a line, adding a sense of movement and liveliness to the scene. The birds' positions vary, with some being closer to the water and others higher up in the sky. The overall atmosphere of the image suggests a pleasant, active day at the beach.
\\ \midrule
Gemini-Rating for Response 2  & Helpfulness:3, Correctness 3, Coherence: 3, Complexity: 2, Verbosity: 3. & Helpfulness:4, Correctness 4, Coherence: 4, Complexity: 3, Verbosity: 3.
\\ \bottomrule
\end{tabular}
}
\vspace{1mm}
\vspace{-5mm}
\label{tab:example_bench}  
  \end{minipage}
\end{table}
\clearpage
\begin{table}[t]
\renewcommand\tabcolsep{0.5pt}
  \begin{minipage}{0.99\textwidth}
\centering
\scalebox{0.88}{
\begin{tabular}{l p{5.4cm} p{7.1cm}}
\toprule
\multicolumn{3}{l}{\bf Gemini-Annotation example (continued)} \\
\\ \midrule
Gemini Chain-of-thought Reasoning  & Both responses are relevant to the question and provide good descriptions of the room. However, the first response is more detailed and provides more analysis of the potential artistic or creative significance of the room. I will give the first response a slightly higher rating for Helpfulness, Correctness, and Complexity. & I think that Response 2 is the best response because it is the most detailed, engaging, and accurate. It provides the most information about the image and uses the most vivid language. Response 1 is also good, but it is not as detailed or engaging as Response 2. 
\\ \midrule
LLaVA-RLHF rated preference & Reponse 2 & Response 2
\\ \midrule
Our preference & Response 1 & Response 2
\\ \bottomrule
\end{tabular}
}
\vspace{1mm}
\vspace{-5mm}
  \end{minipage}
\label{tab:AnnotationExample}
\end{table}

\begin{table}[t]
\centering
\caption{Results of DPO hyperparameter search on LLaVA bench.}
\label{tab:DPO-hyperparam}
\begin{tabular}{cccc}
\toprule
\textbf{Beta Value} & \textbf{Average Log Probability} & \textbf{Learning Rate} & \textbf{LLaVA-bench } \\
\midrule
0.1 & True  & 5e-5 & 60.4 \\
0.1 & False & 5e-5 & 77.4 \\
0.1 & True  & 5e-6 & 73.8 \\
0.1 & False & 5e-6 & 74.4 \\
0.5 & True  & 5e-5 & 59.1 \\
0.5 & False & 5e-5 & 75.3 \\
0.5 & True  & 5e-6 & 74.1 \\
0.5 & False & 5e-6 & 73.1 \\
\bottomrule
\end{tabular}
\end{table}

\begin{figure}[H]
  \centering
  \includegraphics[width=0.6\textwidth]{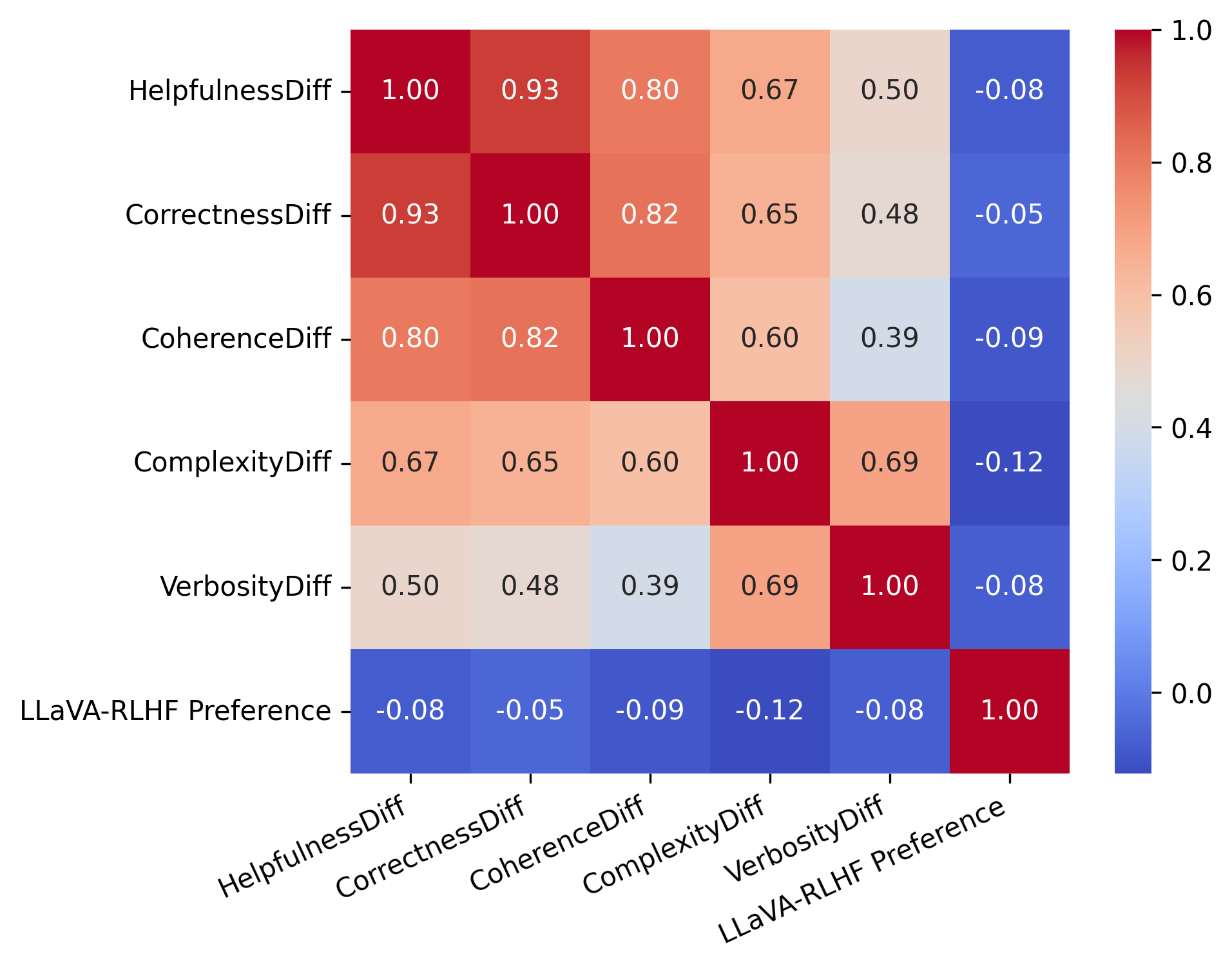}
  \caption{Pearson Correlation Heatmap among the difference in Gemini-Annotated data attributes and LLaVA-RLHF human annotated preference (n=500).}
  \label{fig:correlation-heatmap}
\end{figure}
\end{document}